\documentclass[sigconf]{acmart}

\AtBeginDocument{%
  \providecommand\BibTeX{{%
    \normalfont B\kern-0.5em{\scshape i\kern-0.25em b}\kern-0.8em\TeX}}}

\copyrightyear{2021}
\acmYear{2021}
\setcopyright{acmcopyright}\acmConference[WSDM '21]{Proceedings of the Fourteenth ACM International Conference on Web Search and Data Mining}{March 8--12, 2021}{Virtual Event, Israel}
\acmBooktitle{Proceedings of the Fourteenth ACM International Conference on Web Search and Data Mining (WSDM '21), March 8--12, 2021, Virtual Event, Israel}
\acmPrice{15.00}

\usepackage{enumitem}
\usepackage{balance}
\usepackage{amsmath,amsfonts}
\usepackage{algorithmic}
\usepackage{graphicx}
\usepackage{textcomp}
\usepackage{xcolor}
\def\BibTeX{{\rm B\kern-.05em{\sc i\kern-.025em b}\kern-.08em
    T\kern-.1667em\lower.7ex\hbox{E}\kern-.125emX}}
    
\usepackage{booktabs}
\usepackage[normalem]{ulem}
\useunder{\uline}{\ul}{}

\usepackage{changepage}
\usepackage{pgfplots}

\usepackage{subcaption}

\usepackage{mathtools} 
\usepackage[font=normalsize]{caption}
\pgfplotsset{compat=1.16}
\usepackage{fancyhdr}

\let\oldmaketitle\maketitle
\renewcommand{\maketitle}{%
  \oldmaketitle%
  \thispagestyle{fancy}}
  
\begin{document}

\pagestyle{fancy}
\fancyhf{}

\title{Origin-Aware Next Destination Recommendation with Personalized Preference Attention}


\author{Nicholas Lim}
\affiliation{GrabTaxi Holdings, Singapore}
\email{nic.lim@grab.com}

\author{Bryan Hooi}
\affiliation{Institute of Data Science and School of Computing, National University of Singapore, Singapore}
\email{dcsbhk@nus.edu.sg}

\author{See-Kiong Ng}
\affiliation{Institute of Data Science and School of Computing, National University of Singapore, Singapore}
\email{seekiong@nus.edu.sg}

\author{Xueou Wang}
\affiliation{Grab-NUS AI Lab, National University of Singapore, Singapore}
\email{idswx@nus.edu.sg}

\author{Yong Liang Goh}
\affiliation{GrabTaxi Holdings, Singapore}
\email{yongliang.goh@grab.com}

\author{Renrong Weng}
\affiliation{GrabTaxi Holdings, Singapore}
\email{renrong.weng@grab.com}

\author{Rui Tan}
\affiliation{GrabTaxi Holdings, Singapore}
\email{rui.tan@grab.com}

\begin{abstract}
Next destination recommendation is an important task in the transportation domain of taxi and ride-hailing services, where users are recommended with personalized destinations given their current origin location. However, recent recommendation works do not satisfy this origin-awareness property, and only consider learning from historical destination locations, without origin information. Thus, the resulting approaches are unable to learn and predict origin-aware recommendations based on the user's current location, leading to sub-optimal performance and poor real-world practicality. Hence, in this work, we study the origin-aware next destination recommendation task. We propose the Spatial-Temporal Origin-Destination Personalized Preference Attention (STOD-PPA) encoder-decoder model to learn origin-origin (OO), destination-destination (DD), and origin-destination (OD) relationships by first encoding both origin and destination sequences with spatial and temporal factors in local and global views, then decoding them through personalized preference attention to predict the next destination. Experimental results on seven real-world user trajectory taxi datasets show that our model significantly outperforms baseline and state-of-the-art methods.
\end{abstract}

\ccsdesc[500]{Information systems~Recommender systems}

\keywords{Recommender System; Recurrent Neural Network; Spatio-Temporal}

\maketitle

\section{Introduction}

Recent years have seen rapid growth in the popularity of ride-hailing and mobile applications for taxi-booking, such as Uber, Didi, and others, where users book taxi rides from an origin (O) to a destination (D) location. This surge in taxi booking transactions has resulted in massive user trajectory datasets that provide the opportunity to learn to perform personalized recommendations for users, such as predicting their next destination location.

Different from the popular existing taxi trajectory datasets \cite{MPE,TTDM} that record the sequence of locations from multiple different customers or users for the same taxi, a user trajectory instead records the taxi riding patterns for the \emph{same user}. Thus, they record the sequence of past taxi trips by a user, in the form of origin-destination (OD) tuples that reflect her transport mobility behaviors and latent preferences. Accordingly, these user trajectory datasets present the opportunity to learn each user's preferences, in order to provide personalized recommendations, and to improve performance on the next destination recommendation task.

Existing works for the next Point-of-Interest (POI) or destination location recommendation task mainly focus on the Location-Based Social Network (LBSN) domain, where they learn from LBSN's user trajectories datasets, containing only users' sequential historical destinations or checked-in locations with \emph{no origin information} to predict the next destination. However, in the transportation domain, the historical taxi trips do not just record where the user has visited (i.e. destination), but also where the user has come from (i.e. origin). The inclusion of such origin information can provide several key benefits to existing next destination recommendation works that currently only rely on destination sequences. First, as shown in Fig. 1, the sequential transition of destinations from $d_{t_{1}}$ to $d_{t_{2}}$ can be used to learn destination-destination (DD) relationships and predict the next destination of $d_{t_{i}}$, as studied extensively in existing recommendation works. However, we can also see that the associating origin locations of $o_{t_{1}}$, $o_{t_{2}}$ and current origin $o_{t_{i}}$ can serve as a valuable source of input to learn both origin-origin (OO) and OD relationships to help predict the next destination $d_{t_{i}}$ correctly, but such relationships (dashed arrows on Fig. 1) have not been studied by existing works. Second, by conditioning the next destination prediction $d_{t_{i}}$ on the current origin $o_{t_{i}}$ or where the user is currently at when booking a taxi ride, this extends the problem to \emph{origin-aware next destination recommendation}, where this conditioning of origin-awareness eliminates the problem whereby the same recommendation set is always suggested regardless of where the user is, which is impractical in a real-world setting (e.g. very far away destinations should not be recommended).

Among the recent existing works which do not consider origin information for their next destination recommendations, RNN-based approaches have been shown to surpass classical methods of Markov Chains (MCs) and Matrix Factorization (MF). For instance, \cite{strnn} proposed the Spatial Temporal Recurrent Neural Network (ST-RNN) to exploit spatial and temporal intervals between neighbouring destination locations,  where a time window is used to consider several destination locations as input. \cite{hstlstm} proposed the Hierarchical Spatial-Temporal Long-Short Term Memory (HST-LSTM) to leverage spatial and temporal intervals directly into the existing multiplicative gates of LSTM. \cite{stgcn} proposed the Spatio-Temporal Gated Coupled Network (STGCN) to capture short and long-term user preferences with new distance and time gates from the continuous spatial and temporal intervals. \cite{LSTPM} proposed the Long- and Short-Term Preference Modeling (LSTPM) to learn long and short term user preferences through a nonlocal network and a geo-dilated RNN respectively and is the state-of-the-art approach.


In this paper, we study the origin-aware next destination recommendation task by proposing a Spatial-Temporal Origin-Destination Personalized Preference Attention (STOD-PPA) model based on an encoder-decoder framework to learn the OO, DD, and OD relationships as shown in Fig. 1. The STOD-PPA model first encodes both historical origin and destination sequences using a novel Spatial-Temporal LSTM (ST-LSTM) module, an extension of the LSTM to enable OD relationships to be learned by new cell states based on spatial and temporal factors from both local and global views. Unlike recent works that focus on short and long term user preferences \cite{LSTPM,stgcn}, we propose a novel Personalized Preference Attention (PPA) decoder module to compute a hidden representation for the prediction task by performing personalized preference attention directly on all encoded hidden states of both OD sequences to best learn the dynamic preferences of the user.
To summarise, the contributions of this paper are as follows:
\begin{itemize}[leftmargin=*,topsep=0pt]
\item We propose a novel encoder-decoder STOD-PPA model to learn OO, DD, and OD relationships for the origin-aware next destination recommendation task.  To the best of our knowledge, we are the first work to study the origin-aware next destination recommendation task.

\item Our proposed ST-LSTM encoder module is the first to consider both local and global views for spatial and temporal factors and incorporate them into their own spatial and temporal cell states to learn OD relationships.
\item Experiments conducted on seven real-world user trajectory taxi datasets of different countries in the transportation domain show that our approach outperforms baseline and state-of-the-art methods significantly.
\end{itemize}

\raggedbottom

\begin{figure}  [t]
  \centering
  \includegraphics[width=0.68\linewidth,height=4.3cm]{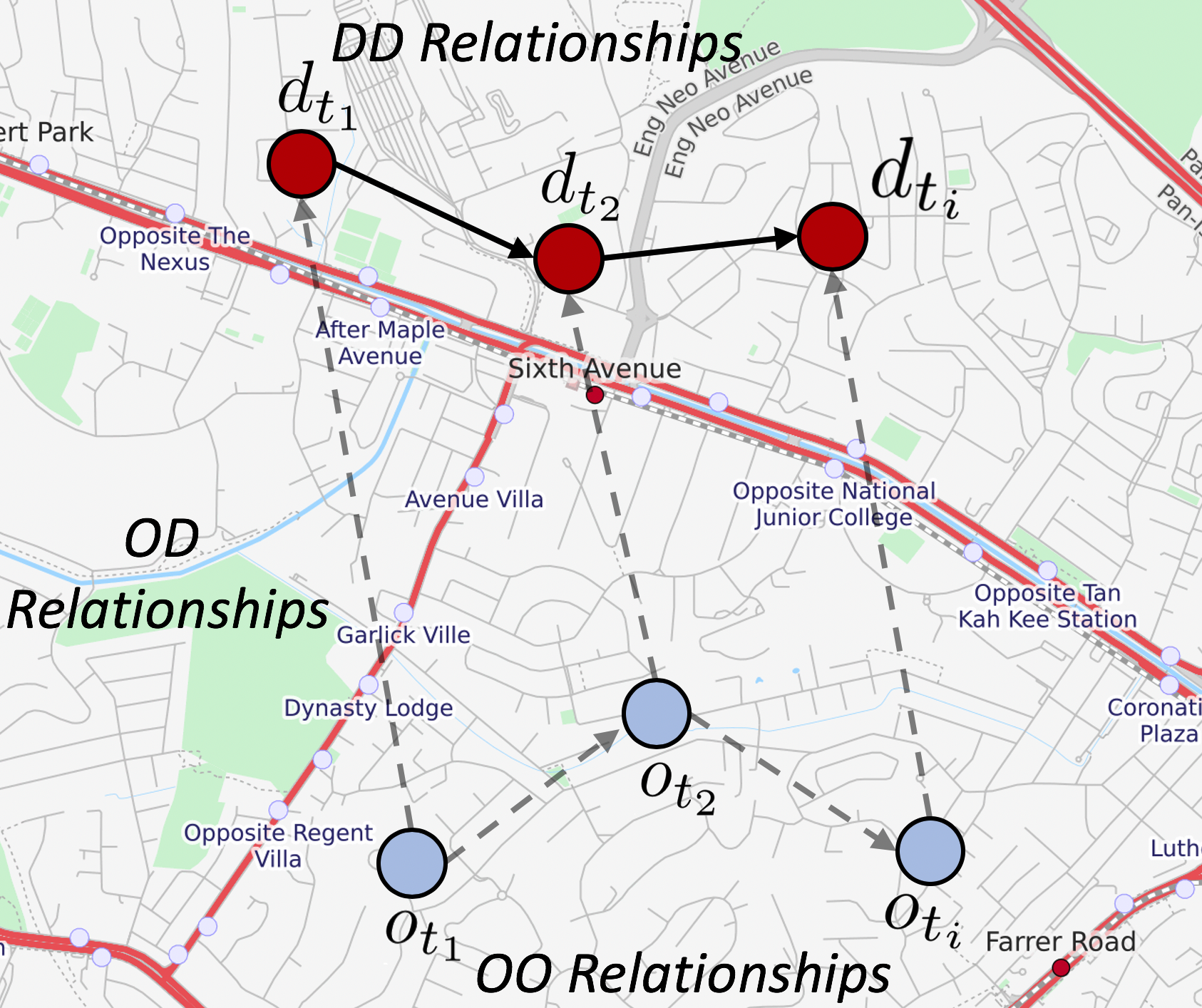}
    \caption{Illustration of a user trajectory containing both origin (blue nodes) and destination (red nodes) sequences, available in the transportation domain to predict the next destination $d_{t_{i}}$. Maps © OpenStreetMap contributors, CC BY-SA.}
\end{figure}

\section{Related Work}
The closest line of work to our origin-aware next destination recommendation task is the next Point-of-Interest (POI) or location recommendation task, which does not use origin information, but only sequentially visited locations (i.e. destinations). This problem is mostly studied in the LBSN domain, such as the popular Gowalla, Brightkite, and Foursquare datasets \cite{gowallaBrightkiteData,sgFS}. Early works studied conventional sequential approaches of MC, as well as collaborative filtering via MF. FPMC-LR \cite{FPMC-LR} extended FPMC \cite{FPMC} to incorporate personalized MC and applied a localized region constraint where only nearby new POIs around the users' historical POIs are considered for the recommendation task, reducing noisy information and improves user experience \cite{treeBased}. PRME \cite{PRME} learns both sequential transitions of POIs and user preferences by modelling them in a latent space. PRME-G is a variant that considers the geographical influence to improve the recommendations. \cite{thirdRankTensor} proposed a Bayesian Personalized Ranking (BPR) approach to learn check-in behaviors and fuses the personalized MC with latent patterns. CAPE \cite{CAPE} learns content-aware POI embeddings from user check-in sequences and POI textual information. \cite{ge} proposed GE, an approach to learn POI, time slot, region, and word embeddings from various graphs, such as POI-POI graphs \cite{unifying,adversarial,ensemble,stpudgat}, then using these embeddings to score each POI for the recommendation task. 

Recently, RNN-based approaches have been demonstrated to perform better than these existing methods due to their ability to model sequential dependencies in user location sequences. ST-RNN \cite{strnn} first proposed the use of spatial and temporal intervals among sequences of POIs by incorporating them into an RNN after performing linear interpolation. \cite{hstlstm} proposed HST-LSTM, a hierarchical LSTM model for session data, and similarly incorporates spatial and temporal intervals into LSTM's existing gates after using linear interpolation. STGN \cite{stgcn} is a LSTM-based model that includes new time and distance gates to incorporate temporal and spatial intervals respectively. The model also includes an additional cell state so that both short and long-term user preferences can be learned. STGCN, a variant of STGN, was also proposed where the forget gate is removed for efficiency. CatDM \cite{categoryAware} considers both POI categories and spatial proximity to mitigate data sparsity, but is limited to only predicting the POIs visited in the next window of 24 hours. DeepMove \cite{deepMove} uses a historical attention module and a GRU to learn sequential transitions from visit sequences. LSTPM \cite{LSTPM} learns both long and short term user preferences through a nonlocal network architecture where the spatial and temporal correlations between current and past trajectories are considered. The long term preferences exploit the historical trajectories using a nonlocal network, while the short term preferences focus on the most recent trajectory to model users' latest preferences using a geo-dilated RNN. LSTPM is also the state-of-the-art model for the next POI recommendation task. However, these existing works are not designed to work with origin information.

Among the other existing works, approaches that consider the use of both origin and destination information are limited, where the only work of \cite{grabPOI} applied a simple frequency approach under their suggestion framework to compute a recommendation list by considering both user's historical destinations and popular city-wide locations; however, this approach is \emph{not origin-aware}. In this paper, we only consider their suggestion framework as their prediction framework does not compute a ranked list for evaluation.



\section{Preliminaries}

\paragraph{\textbf{Problem Formulation}} Let  $U=\{u_{1},u_{2},...,u_{M}\}$ be the set of $M$ users and $L=\{l_{1},l_{2},...,l_{N}\}$
be the set of $N$ locations for the users in $U$ to visit, where each location $l_{n} \in L$ can either have the role of an origin $o$ or a destination $d$, or both among the dataset of user trajectories. Each user $u_{m}$ has an OD sequence of pick-up (origin) and drop-off (destination) tuples $s_{u_{m}}=\{(o_{t_{1}},d_{t_{1}}),(o_{t_{2}},d_{t_{2}}),...,(o_{t_{i}},d_{t_{i}})\}$ that record their taxi trips, and $S=\{s_{u_{1}},s_{u_{2}},...,s_{u_{M}}\}$ is the set of all users' OD sequences. All location visits in $S$, regardless of origin or destination, have their own location coordinates $LC$ and timestamp $time$.  The objective of the origin-aware next destination recommendation task is to consider the user $u_{m}$, the current origin $o_{t_{i}}$, and her historical sequence of OD tuples $\{(o_{t_{1}},d_{t_{1}}),(o_{t_{2}},d_{t_{2}}),...,(o_{t_{i-1}},d_{t_{i-1}})\}$ to recommend an ordered set of destinations from $L$, where the next destination $d_{t_{i}}$ should be highly ranked in the recommendation set. We further denote $s_{u_{m}}^{train}$ and $S^{train}$ as sets from the training partition with the superscript $train$ for clarity.

\subsection{LSTM}
The LSTM \cite{lstm} is a variant of RNN that is capable of learning sequential transitions across timesteps of the sequence through gating mechanisms and a memory cell state to retain information. The gating mechanism helps to mitigate the vanishing gradient problem for sequences and has been found effective in various sequential predictive applications. Here, we include the original LSTM model and its equations:
\begin{gather}
i_{t_{i}} = \sigma(\textbf{W}_{i} \: x_{t_{i}}  + \textbf{U}_{i} \:  h_{t_{i-1}} + \textbf{b}_{i})
\\
f_{t_{i}} = \sigma(\textbf{W}_{f} \: x_{t_{i}} + \textbf{U}_{f} \:  h_{t_{i-1}} + \textbf{b}_{f})
\\
o_{t_{i}} = \sigma(\textbf{W}_{o} \: x_{t_{i}}  + \textbf{U}_{o} \:  h_{t_{i-1}} + \textbf{b}_{o})
\\
\Tilde{c}_{t_{i}} = tanh(\textbf{W}_{c} \:  x_{t_{i}} + \textbf{U}_{c} \:  h_{t_{i-1}} + \textbf{b}_{c})
\\
c_{t_{i}} = f_{t_{i}} \odot c_{t_{i-1}} + i_{t_{i}} \odot \Tilde{c}_{t_{i}}
\\
h_{t_{i}} = o_{t_{i}} \odot tanh(c_{t_{i}})
\end{gather}
where $i_{t_{i}}$, $f_{t_{i}}$, $o_{t_{i}}$ are the input, forget and output gates respectively that regulate information flow in the scale of 0 to 1 from the sigmoid activation function. For each timestep $t_{i}$, the LSTM extracts information via the input gate $i_{t_{i}}$ from the cell input $\Tilde{c}_{t_{i}}$ using Hadamard product $\odot$, representing ``how much to store in the cell state $c_{t_{i}}$''. The forget gate $f_{t_{i}}$ regulates ``how much to forget from the previous cell state $c_{t_{i-1}}$''. The LSTM outputs $c_{t_{i}}$ and $h_{t_{i}}$ for the next timestep accordingly, where $h_{t_{i}}$ is the hidden state or output hidden representation regulated by the output gate indicating ``how much to output for this timestep'' from the current cell state $c_{t_{i}}$.

\section{Approach}
In this section, we first propose the ST-LSTM module (encoder), an extension of the LSTM to learn OD relationships, followed by the PPA module (decoder) to best learn users' dynamic preferences.

\subsection{ST-LSTM} \label{sec:stlstm}


Recent RNN and LSTM based works \cite{stgcn,hstlstm,strnn} have incorporated spatial (distance) and temporal (time) intervals of $\Delta d_{t_{i}}$ and $\Delta t_{t_{i}}$ respectively, between neighbouring pairs of destinations as input to their models. For example, given the previous destination $d_{t_{i-1}}$ and the next destination $d_{t_{i}}$ to be visited and predicted by the model, the spatial interval $\Delta d_{t_{i}} = d(LC_{t_{i}},LC_{t_{i-1}})$ is computed using a distance function $d$ on the location coordinates $LC$ of both destinations $d_{t_{i}}$ and $d_{t_{i-1}}$. Similarly, the temporal interval $\Delta t_{t_{i}} = time_{t_{i}} - time_{t_{i-1}} $ is computed from the corresponding timestamps $time$ of both $d_{t_{i}}$ and $d_{t_{i-1}}$ destinations. Intuitively, the spatial and temporal intervals seek to describe ``how far is the next destination'' and ``how long before they visit the next destination'' respectively, using these intervals as inputs to their models accordingly to predict the next destination $d_{t_{i}}$. Notably, this usage of intervals makes the key assumption that the next destination $d_{t_{i}}$ and its details (i.e. location coordinates $LC_{t_{i}}$ and timestamp $time_{t_{i}}$) are known in advance, making it impractical in a real-world setting where the next destination visit $d_{t_{i}}$ is not yet known \cite{deepMove}. Apart from this impracticality flaw, the consideration of spatial and temporal factors between only neighbouring destination pairs restricts the model to only learn from a local view and not from a global view where all locations from $L$ are involved.

Inspired from these two flaws, we propose to extend the LSTM with spatial and temporal cell states to learn OD relationships based on the spatial and temporal factors in both local and global views. This extension seeks to allow OD relationships to be learned as the LSTM is already capable of learning OO or DD relationships given an origin or a destination sequence respectively but is unable to learn OD relationships.

\paragraph{\textbf{Learning Spatial OD Relationships}} We propose the use of Geohash embeddings (local view) and spatial intervals (global view) to learn spatial OD relationships among all locations in $L$, which includes both origins and destinations. First, we partition the map with a Geohash grid.
As each location $l_{n} \in L$ has its own location coordinates $LC$, we map $l_{n}$ to its corresponding Geohash cell $l_{n}^{geo}$, as well as the Geohash embedding of that cell $\vec{l}_{n}^{geo}$. Accordingly, this means that locations in the same region or Geohash would be trained to be similar to each other under a \emph{local view}, such as the origin $l_1$ and destinations of $l_2$ and $l_3$ in Fig. 2(a), allowing region specific semantics such as shopping districts and housing neighbourhood areas to be learned. However, a limitation with this approach is that the relationship between $l_1$ and $l_4$ may be deemed dissimilar by the model due to the assignment of different Geohash embeddings based on their respective Geohash cells, even when they are near to each other, due to their different cells or areas.

To mitigate this limitation, we propose to also learn the spatial factor from a global view. In Fig. 2(b), we can see that instead of partitioning the same map and locations by a Geohash grid, for the global view, we compute pair-wise spatial or distance intervals $\Delta s$ between the origin $l_{1}$ and all locations available of $\{l_1,l_2,l_3,l_4,l_5\}$ (including $l_{1}$ itself for simplicity). The intervals would be able to embed the spatial proximity of the current location $l_{1}$ to all the other locations and itself, where the origin $l_{1}$ can now learn that it is also similar or nearby to the destination $l_{4}$, which was not achievable under a local view as shown on Fig. 2(a). However, the global view also has a limitation of being unable to learn area or Geohash specific semantics, necessitating the consideration of both local and global views to learn OD relationships.


\begin{figure}[t]
\begin{subfigure}[t]{0.5\linewidth-4pt}
  \centering
  \includegraphics[width=1\linewidth,height=3.5cm]{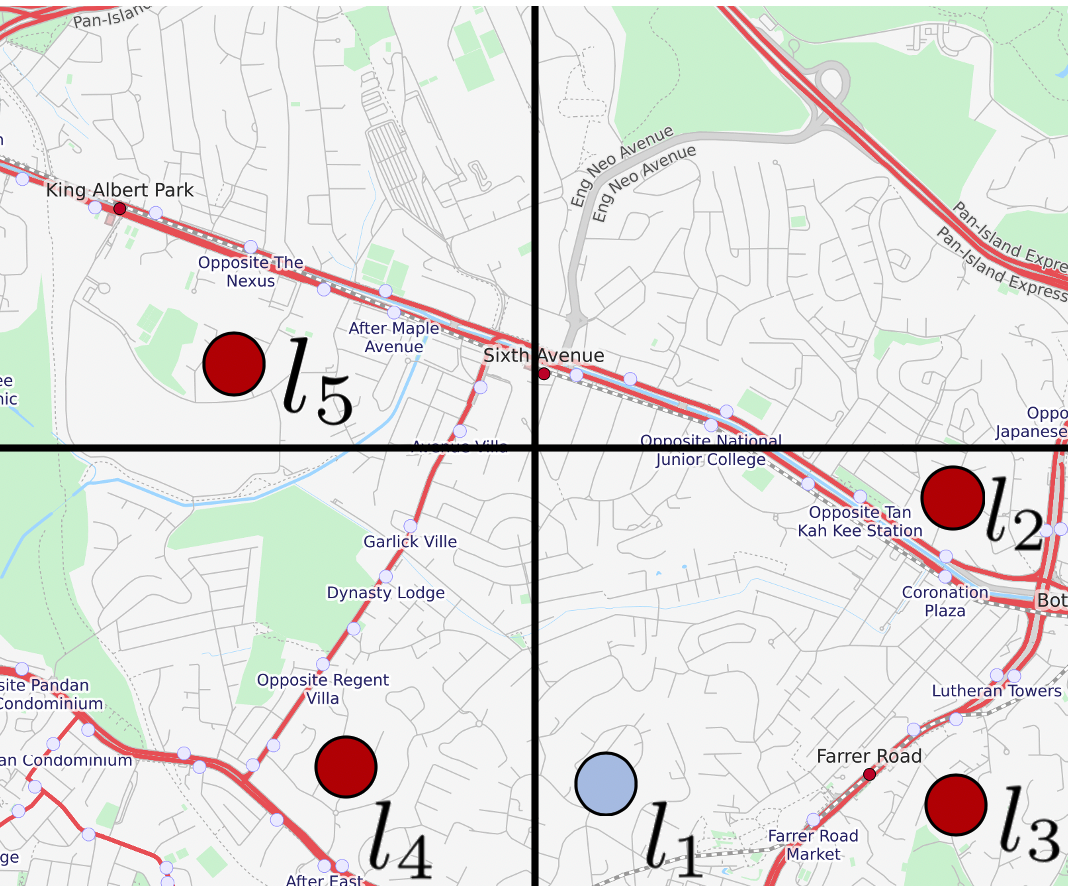}  
  \caption{\emph{Local View}: Learning OD similarities from Geohash cells.}
  \label{fig:sub-first}
\end{subfigure}%
~\hspace{6pt}
\begin{subfigure}[t]{0.5\linewidth-4pt}
  \centering
  \includegraphics[width=1\linewidth,height=3.5cm]{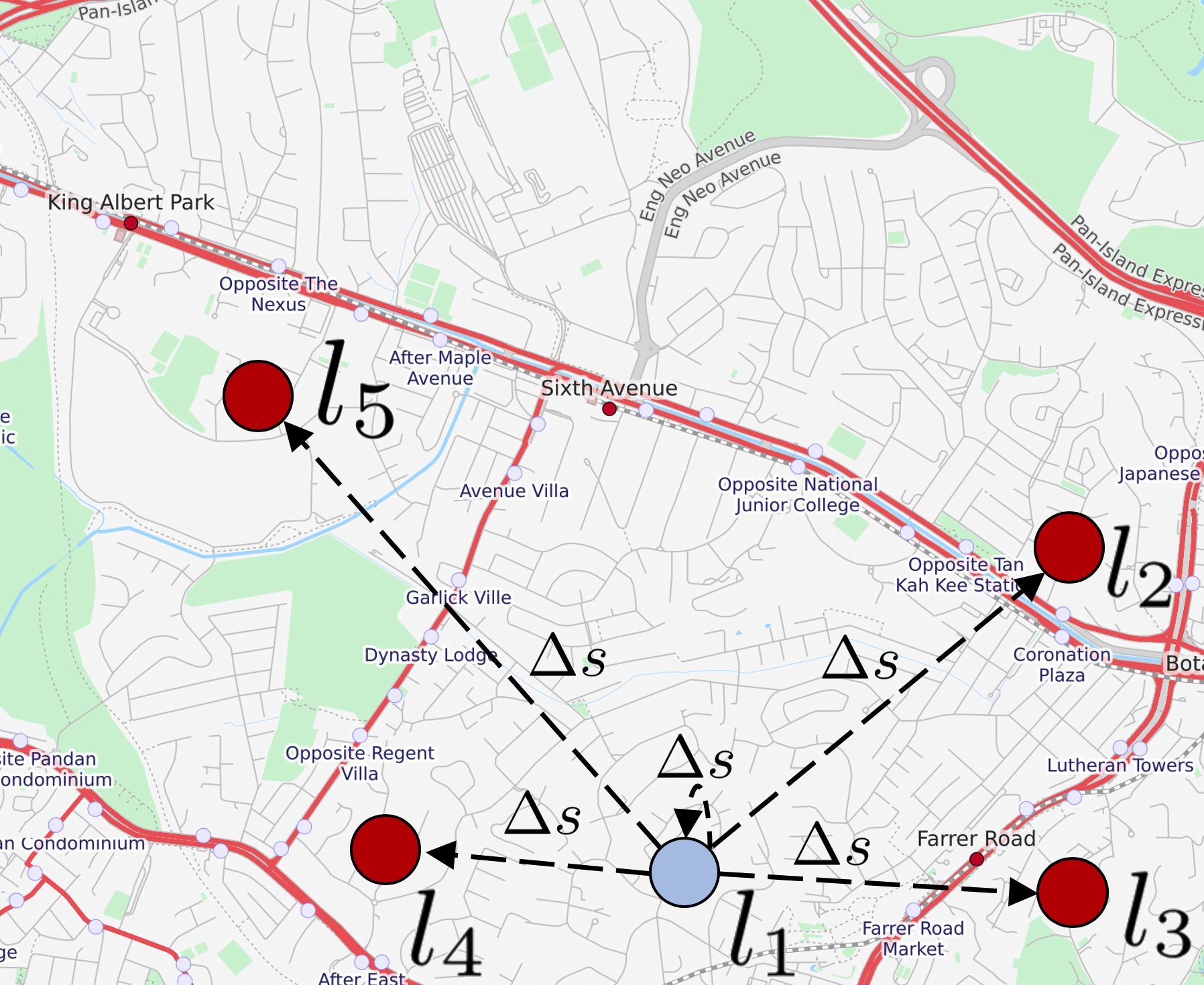}  
  \caption{\emph{Global View}: Learning OD similarities from spatial intervals.}
  \label{fig:sub-second}
\end{subfigure} 
\hspace{2pt}
\vspace{-10pt}
\caption{Learning spatial factors of origins (blue nodes) and destinations (red nodes). Maps © OpenStreetMap contributors, CC BY-SA.}
\label{fig:fig}
\vspace{-0.3cm}
\end{figure}

Now, instead of just considering $\{l_1,l_2,l_3,l_4,l_5\}$ for the example in Fig. 2(b), we perform the pair-wise spatial interval computation between $l_1$ and all locations $l_{n} \in L$ using the Haversine distance function, to compute a vector representation of the spatial intervals $\Delta \vec{s}_{l_{1}} \in \mathbb{R}^{|L|} $ for $l_{1}$ to represent the \emph{global view}. Intuitively, this provides a global context on how far or how near are all other locations (origin or destination) in the whole city or country is to $l_{1}$, whereas the local view focuses on region level semantics. Different from the existing works that require the last spatial and temporal intervals of $\Delta d_{t_{i}}$ and $\Delta t_{t_{i}}$ respectively as input to the model \cite{strnn,hstlstm,stgcn} to predict the next destination $d_{t_{i}}$, here, we compute intervals between a location $l_1$ to all locations $l_n \in L$ without the need to know the next destination location $d_{t_{i}}$ in advance.

With an input location embedding $\vec{l}_{t_{i}} \in \mathbb{R}^{dim}$ as $x_{t_{i}}$ for Eq. (1) to (4) of the LSTM, as well as its mapped Geohash embedding $\vec{l}^{geo}_{t_{i}} \in \mathbb{R}^{dim}$ (local view) and its spatial interval vector $\Delta \vec{s}_{l_{t_{i}}} \in \mathbb{R}^{|L|}$ (global view), where $dim$ is the embedding dimension, we propose a new spatial cell state and its associating gates and cell input \emph{in addition} to LSTM's cell state: 
\begin{gather}
i_{t_{i}}^{s} = \sigma(\textbf{W}_{i}^{s} \: \vec{l}^{geo}_{t_{i}} + \textbf{V}_{i}^{s}\: \Delta \vec{s}_{l_{t_{i}}} + \textbf{U}_{i}^{s} \:  h_{t_{i-1}} + \textbf{b}_{i}^{s})
\\
f_{t_{i}}^{s} = \sigma(\textbf{W}_{f}^{s} \: \vec{l}^{geo}_{t_{i}} + \textbf{V}_{f}^{s}\: \Delta \vec{s}_{l_{t_{i}}} + \textbf{U}_{f}^{s} \:  h_{t_{i-1}} + \textbf{b}_{f}^{s})
\\
\Tilde{c}_{t_{i}}^{s} = tanh(\textbf{W}_{c}^{s} \: \vec{l}^{geo}_{t_{i}} + \textbf{V}_{c}^{s}\: \Delta \vec{s}_{l_{t_{i}}}  + \textbf{U}_{c}^{s} \: h_{t_{i-1}} + \textbf{b}_{c}^{s})
\\
c_{t_{i}}^{s} = f_{t_{i}}^{s} \odot c_{t_{i-1}}^{s} + i_{t_{i}}^{s} \odot \Tilde{c}_{t_{i}}^{s}
\end{gather}
where $i_{t_{i}}^{s}, f_{t_{i}}^{s}, \Tilde{c}_{t_{i}}^{s}, c_{t_{i}}^{s}$ are the set of input gate, forget gate, cell input and cell state respectively to learn the sequential spatial transitions with the superscript $s$. Here, $\textbf{W}_{i}^{s} ,\textbf{W}_{f}^{s},\textbf{W}_{c}^{s} \in \mathbb{R}^{dim \times Hdim}$ learns a projection of the Geohash embedding $\vec{l}^{geo}_{t_{i}}$ (local view) for the gates $i_{t_{i}}^{s}, f_{t_{i}}^{s}$ and cell input $\Tilde{c}_{t_{i}}^{s}$ respectively, where $Hdim$ is the hidden representation dimension. Similarly, for the global view, $\textbf{V}_{i}^{s},\textbf{V}_{f}^{s},\textbf{V}_{c}^{s} \in \mathbb{R}^{|L| \times Hdim}$ learns a representation from the spatial interval vector $\Delta \vec{s}_{l_{t_{i}}}$. Same as Eq. (1) to (4), we also add $\textbf{U}_{i}^{s},\textbf{U}_{f}^{s},\textbf{U}_{c}^{s} \in \mathbb{R}^{Hdim \times Hdim}$ and their biases $\textbf{b}_{i}^{s},\textbf{b}_{f}^{s},\textbf{b}_{c}^{s} \in \mathbb{R}^{Hdim}$ for the previous hidden state $h_{t_{i-1}}$, as this enforces the recurrent structure where the output hidden state of the previous timestep is used as input to the current timestep to model the sequential spatial transitions. The spatial cell state $ c_{t_{i}}^{s}$ is then computed from Eq. (10) from its own set of gates and cell input, encoding the sequential spatial factor for use by the next timesteps accordingly.

\begin{figure}[t]
  \centering
  \includegraphics[width=0.95\linewidth,height=4.5cm]{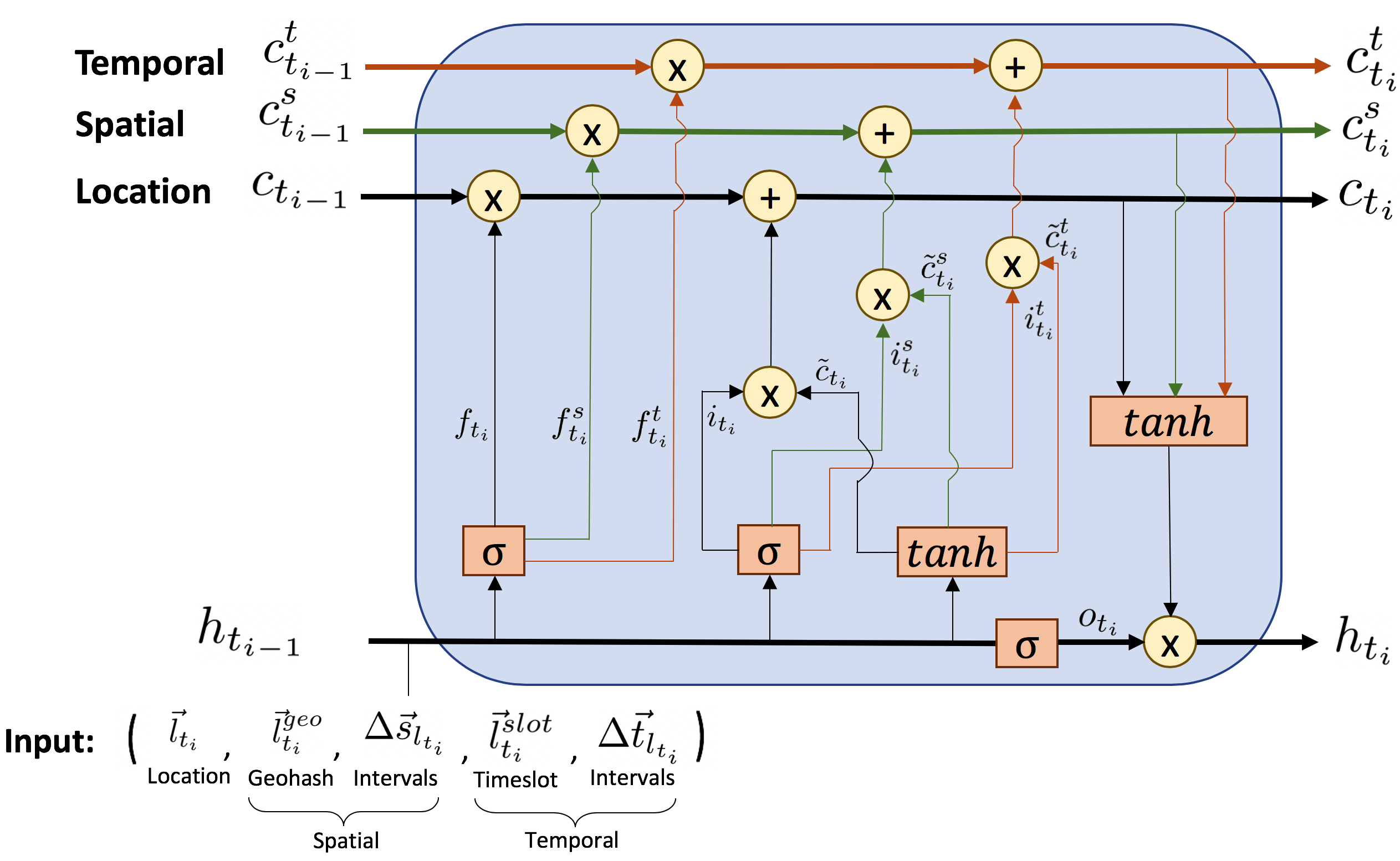}
  \vspace{-0.3cm}
    \caption{ST-LSTM module where green and brown connections are the proposed spatial and temporal cell states.}
    \vspace{-0.3cm}
    
\end{figure}
\paragraph{\textbf{Learning Temporal OD Relationships}} Similar to the spatial cell state, we also propose a \emph{separate} temporal cell state $c_{t_{i}}^{t}$ to learn temporal OD relationships from both local and global views: 
\begin{gather}
i_{t_{i}}^{t} = \sigma(\textbf{W}_{i}^{t} \: \vec{l}^{slot}_{t_{i}} + \textbf{V}_{i}^{t}\: \Delta \vec{t}_{l_{t_{i}}} + \textbf{U}_{i}^{t} \:  h_{t_{i-1}} + \textbf{b}_{i}^{t})
\\
f_{t_{i}}^{t} = \sigma(\textbf{W}_{f}^{t} \: \vec{l}^{slot}_{t_{i}} + \textbf{V}_{f}^{t}\: \Delta \vec{t}_{l_{t_{i}}} + \textbf{U}_{f}^{t} \:  h_{t_{i-1}} + \textbf{b}_{f}^{t})
\\
\Tilde{c}_{t_{i}}^{t} = tanh(\textbf{W}_{c}^{t} \: \vec{l}^{slot}_{t_{i}} + \textbf{V}_{c}^{t}\: \Delta \vec{t}_{l_{t_{i}}} + \textbf{U}_{c}^{t} \:  h_{t_{i-1}} + \textbf{b}_{c}^{t})
\\
c_{t_{i}}^{t} = f_{t_{i}}^{t} \odot c_{t_{i-1}}^{t} + i_{t_{i}}^{t} \odot \Tilde{c}_{t_{i}}^{t}
\end{gather}
where the main differences with the spatial cell state equations from Eq. (7) to (10), apart from its own temporal weight matrices, are the inclusion of $\vec{l}^{slot}_{t_{i}}\in \mathbb{R}^{dim}$ as the corresponding timeslot embedding of the input location $l_{t_{i}}$ (local view) and $\Delta \vec{t}_{l_{t_{i}}} \in \mathbb{R}^{|L|}$ as the temporal interval vector between the input location $l_{t_{i}}$ and all locations in $L$ (global view). As each location visit (origin or destination) includes timestamp data $time$ (taxi pick-up timestamp for origin and drop-off timestamp for destination), we can map each location visit in $S^{train}$ to its corresponding timeslot $l^{slot}_{t_{i}}$. Similar to learning OD relationships in a Geohash cell for the spatial factor, this would allow origin and destination locations in the same timeslot to learn to be similar under a local view for the temporal factor. We use a total of eight timeslots of three hours each to represent the periodic changes across a day. For the global view of temporal interval vector $\Delta \vec{t}_{l_{t_{i}}}$, we first identify OD tuples in all users' sequences $S^{train}$ that contain $l_{t_{i}}$ (origin or destination), and compute the pair-wise temporal interval as $\Delta t = time^{d} - time^{o} $, where $time^{o}$ and $time^{d}$ are the origin and destination timestamps respectively of the OD tuple. 

Next, we modify Eq. (6) to combine the spatial, temporal and LSTM's memory cell states as the output hidden state for the current timestep:
\begin{gather}
h_{t_{i}} = o_{t_{i}} \odot tanh(\textbf{W}_{h} (c_{t_{i}}\:||\:c_{t_{i}}^{s}\:||\:c_{t_{i}}^{t}))
\end{gather}
where $\textbf{W}_{h} \in \mathbb{R}^{3 \cdot Hdim \times Hdim}$ fuses the three cell states after the concatenate operation $||$, followed by the hyperbolic tangent activation function, then applies the LSTM's existing output gate $o_{t_{i}}$ to learn ``how much to output'' from the fused sequential location, spatial and temporal cell states. Accordingly, we illustrate the ST-LSTM module on Fig. 3.

\subsection{STOD-PPA}
We further propose STOD-PPA based on an encoder-decoder framework to first encode OD sequences using our ST-LSTMs, then use a PPA decoder to perform personalized preference attention to all the encoded OD hidden states.

\paragraph{\textbf{Encoder}}
As each user's sequence of OD tuples $s_{u_{m}}$ is partitioned into training and testing partitions, here, we use the training partition $s_{u_{m}}^{train}=\{(o_{t_{1}},d_{t_{1}}),(o_{t_{2}},d_{t_{2}}),...,(o_{t_{i}},d_{t_{i}})\}$ and split them into separate origin and destination sequences of $s_{u_{m}}^{train^{O}}=\{o_{t_{2}},o_{t_{3}},...,o_{t_{i}}\}$ and $s_{u_{m}}^{train^{D}}=\{d_{t_{1}},d_{t_{2}},...,d_{t_{i-1}}\}$ respectively. For efficiency, we omitted the first origin $o_{t_{1}}$ from $s_{u_{m}}^{train^{O}}$ and the last destination $d_{t_{i}}$ from $s_{u_{m}}^{train^{D}}$ so that both the encoder and decoder will use the same set of input sequences, allowing batch training to be performed for each user. Then, we propose to encode both $s_{u_{m}}^{train^{O}}$ and $s_{u_{m}}^{train^{D}}$ separately with a ST-LSTM each, allowing OO and DD relationships to be learned in their own ST-LSTM respectively, as well as OD relationships from the newly proposed spatial and temporal cell states. First, we use a multi-modal embedding layer $Emb$ to embed the location $l_{t_{i}}$ (origin or destination), its corresponding Geohash $l^{geo}_{t_{i}}$ and timeslot $l^{slot}_{t_{i}}$:
\begin{gather}
(\vec{l}_{t_{i}}, \vec{l}^{geo}_{t_{i}}, \vec{l}^{slot}_{t_{i}}) = \: \underset{\mathclap{\:\:\:\:\:\:\:\:\textbf{W} \in \{\textbf{W}_{L},\textbf{W}_{G},\textbf{W}_{T}\}}}{Emb_{\:\textbf{W}}}\:(l_{t_{i}}, l^{geo}_{t_{i}}, l^{slot}_{t_{i}})
\end{gather}
where $\textbf{W}_{L} \in \mathbb{R}^{|L|\times dim},\textbf{W}_{G} \in \mathbb{R}^{|G|\times dim},\textbf{W}_{T} \in \mathbb{R}^{|T| \times dim}$ are the location, Geohash and timeslot weight matrices respectively, $|G|$ is the total number of corresponding Geohashes after mapping all locations in $L$, and $|T|$ is the number of timeslots, where we use $|T|=8$ of three hours each. 
Accordingly, at each timestep, the input to an ST-LSTM would have an input tuple of the mapped embeddings and the global interval vectors $(\vec{l}_{t_{i}},\vec{l}^{geo}_{t_{i}}, \Delta \vec{s}_{l_{t_{i}}},\vec{l}^{slot}_{t_{i}}, \Delta \vec{t}_{l_{t_{i}}})$, as shown on Fig. 3. Next, we proceed to encode the origin and destination sequences separately with their own ST-LSTMs, denoted as $\phi^{O}(.)$ and $\phi^{D}(.)$ respectively, specifically:
\begin{gather}
h^{O}_{u_{m}} = \phi^{O}(s_{u_{m}}^{train^{O}})
\\
h^{D}_{u_{m}} = \phi^{D}(s_{u_{m}}^{train^{D}})
\end{gather}
where $h^{O}_{u_{m}}$ and $h^{D}_{u_{m}}$ are sets containing all origin and destination encoded hidden states across the timesteps respectively and $|h^{O}_{u_{m}}| = |h^{D}_{u_{m}}|$. Then, we concatenate both $h^{O}_{u_{m}}$ and $h^{D}_{u_{m}}$ for a final set of all hidden states $h^{OD}_{u_{m}}$ for user $u_{m} \in U$ as the output of the encoder, for use by the decoder in training and testing.

\begin{figure}[t]
  \centering
  \includegraphics[width=0.85\linewidth,height=5cm]{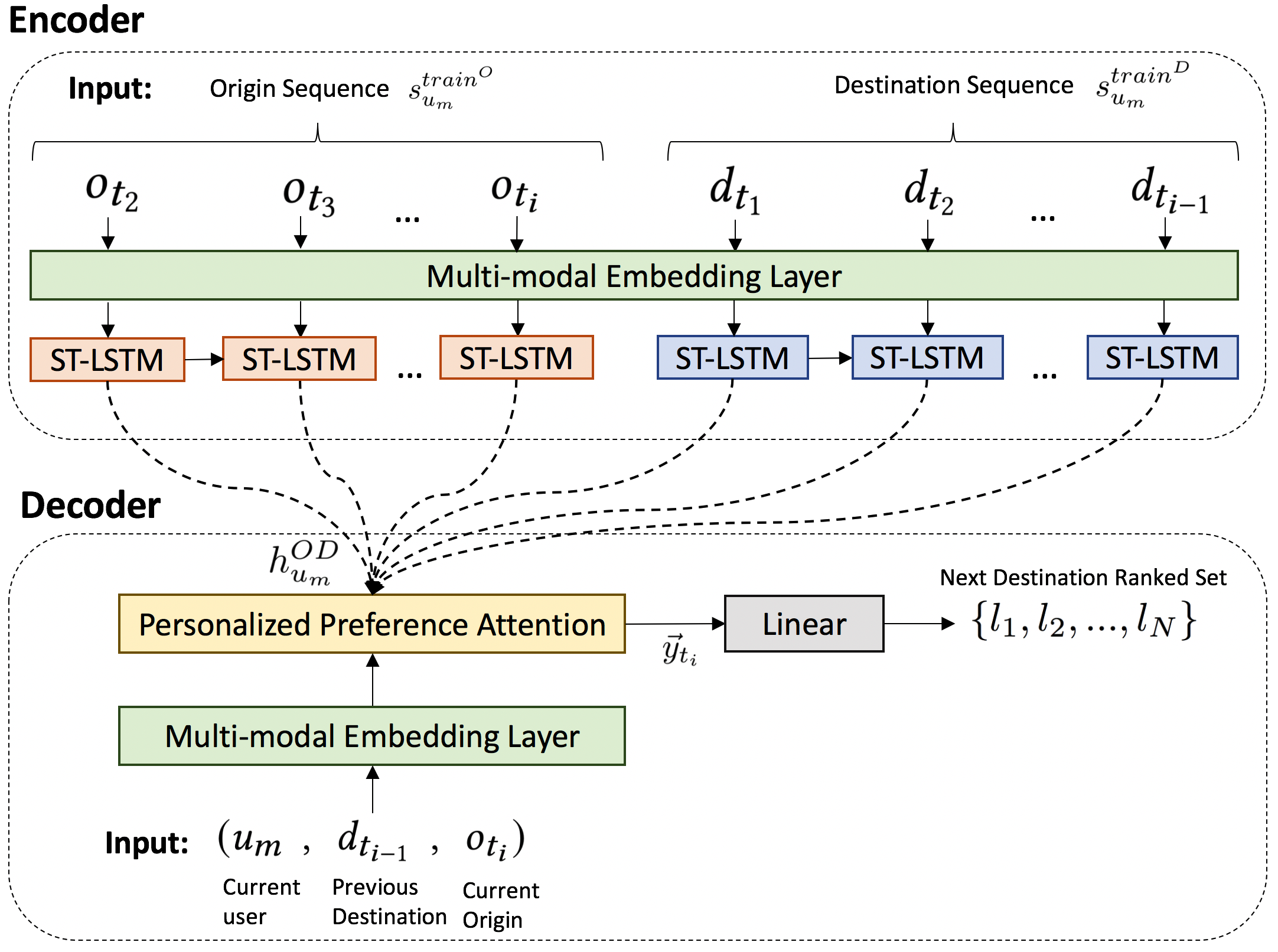}
    \caption{Illustration of STOD-PPA.}
    \vspace{-0.3cm}
\end{figure}

\paragraph{\textbf{Decoder}}
After encoding the OD sequences, we propose the Personalized Preference Attention (PPA) decoder module to attend to all the encoded OD hidden states and compute an origin-aware and personalized hidden representation based on the users' dynamic preferences.
Different from the existing works that includes user embeddings by addition \cite{strnn} or concatenation \cite{deepMove} to the hidden representation used for prediction, here, we propose the novel approach of using user embeddings to learn to perform attention on the encoded OD hidden states. First, we update the multi-modal embedding layer $Emb$ in Eq. (16) with an additional user weight matrix $\textbf{W}_{U} \in \mathbb{R}^{|U| \times dim}$ and use this layer to map user $u_{m}$, current origin $o_{t_{i}}$ and previous destination $d_{t_{i-1}}$ to their corresponding embeddings of $\vec{u}_{m}$, $\vec{o}_{t_{i}}$ and $\vec{d}_{t_{i-1}}$ respectively.
Then, from each user's encoded hidden states $\vec{h}_{i} \in h^{OD}_{u_{m}}$, we can compute its respective personalized attention score:
\begin{gather}
\vec\alpha_{i} = \frac{exp\left(\sigma_{LR}\left(\textbf{W}_{A}(\vec{u}_{m} || \vec{o}_{t_{i}} || \vec{d}_{t_{i-1}} || \vec{h_{i}} )\right)\right)}{\sum_{\vec{p_{i}} \in h^{OD}_{u_{m}}} exp\left(\sigma_{LR}\left(\textbf{W}_{A}(\vec{u}_{m} || \vec{o}_{t_{i}} || \vec{d}_{t_{i-1}} || \vec{p_{i}} )\right)\right)}
\end{gather}
where $\vec\alpha_{i} \in \mathbb{R}^{Hdim}$, $\textbf{W}_{A} \in \mathbb{R}^{(3\cdot dim+Hdim) \times Hdim}$, and $\sigma_{LR}$ is the Leaky ReLU activation function. At every timestep, the decoder takes the input of $\vec{u}_{m}$, $\vec{o}_{t_{i}}$, $\vec{d}_{t_{i-1}}$, and each hidden state $\vec{h}_{i} \in h^{OD}_{u_{m}}$ to predict a personalized attention score $\vec\alpha_{i} \in \alpha_{t_{i}}$ after softmax normalization, where $|\alpha_{t_{i}}| = |h^{OD}_{u_{m}}|$ accordingly. Effectively, the computed personalized attention scores $\alpha_{t_{i}}$ indicates ``how much to attend to each encoded transition or hidden state'' among all the encoded OD hidden states $h^{OD}_{u_{m}}$, that can best predict the next destination $d_{t_{i}}$, by optimizing the trainable weight matrix $\textbf{W}_{A}$.

Unlike recent works \cite{stgcn,LSTPM} that focus on learning long and short term preferences, our approach considers all hidden states directly for the personalized preference attention process to best learn dynamic user preferences. With the personalized attention scores, we then perform a weighted sum on all of the corresponding hidden states:
\begin{gather}
\vec y_{t_{i}} = \sum_{\mathclap{\:\:\:\: \vec\alpha_{i} \in \alpha_{t_{i}},\:\vec{h}_{i} \in h^{OD}_{u_{m}}}}\:\: \vec\alpha_{i} \odot \vec{h}_{i}
\\
P(d_{t_{i}}|o_{t_{i}},d_{t_{i-1}},u_{m}) = softmax(\:\textbf{W}_{loc}(\:\vec y_{t_{i}})\:)
\end{gather}
where $\vec y_{t_{i}} \in \mathbb{R}^{Hdim}$ is the output hidden representation for timestep $t_{i}$. Lastly, we project $\vec y_{t_{i}}$ to the number of locations or $|L|$ where  $\textbf{W}_{loc} \in \mathbb{R}^{Hdim \times |L|} $, followed by a softmax function to derive a probability distribution of all locations by learning $P(d_{t_{i}}|o_{t_{i}},d_{t_{i-1}},u_{m})$ as a multi-classification problem. Accordingly, we can sort the distribution in descending order to achieve the final ranked recommendation set where the next destination location $d_{t_{i}}$ should be highly ranked. Fig. 4 illustrates the STOD-PPA model.

\paragraph{\textbf{Prediction}}
After training STOD-PPA, at only prediction or test time, we \emph{deactivate the encoder} and only use the decoder for prediction as the encoded hidden states $h^{OD}_{u_{m}}$ from each users' historical OD sequences are stored for efficiency after the training phase. The stored or pre-computed $h^{OD}_{u_{m}}$ are then actively retrieved by the decoder at test time for the respective user to perform attention on them and compute a ranked set prediction based on the different test case or input tuple of the user $u_{m}$, current origin $o_{t_{i}}$ and previous destination $d_{t_{i-1}}$, as shown on Fig. 4, in order to compute Eq. (19) to (21).

\section{Experiments}

\subsection{Datasets}

We use seven real-world user trajectory taxi datasets of different Southeast Asia (SE) countries in the recent year of 2019 from the ride-hailing company Grab for evaluation, where users would book taxis from the mobile application. Table 1 shows the details of the datasets (country names omitted to not reflect business insights) where the total number of locations is computed from the total number of origins and destinations, and each trip is an OD tuple accordingly from pick-up (origin) to drop-off (destination). For preprocessing, we use the same settings as \cite{stgcn}, where users with less than 10 OD tuples or trips are removed and locations (i.e. origin or destination) visited by less than 10 users are removed. Lastly, we sort each user's visit records by timestamps in ascending order, taking the first 70\% as the training set and the remaining 30\% as the testing set in order to predict future visited destinations.

\begin{table}[]
\centering
\caption{Statistics of the seven datasets (after preprocessing).}
\resizebox{0.95\linewidth}{1.5cm}{%
\begin{tabular}{@{}cccccc@{}}
\toprule 
Datasets & \#users & \#locations & \#origins & \#destinations & \#trips \\ \midrule
SE-1 & 2,662 & 1,694 & 1,131  & 563  & 23,730 \\
SE-2 & 2,595 & 1,523 & 1,008 & 515 & 22,867 \\
SE-3 & 2,677 & 1,469 & 972 & 497 & 21,164  \\
SE-4 & 3,083 & 1,625 & 1,003 & 622  & 24,344 \\
SE-5 & 2,452  & 1,397  & 891  & 506  & 22,256  \\
SE-6 & 1,363 & 1,001 & 642 & 359 & 21,156 \\ 
SE-7 & 3,301 & 2,044 & 1,315 & 729 & 23,019 \\\bottomrule
\end{tabular}%
}
\vspace{-0.3cm}
\end{table}

\subsection{Baseline Methods and Evaluation Metrics}

\begin{itemize}[leftmargin=*,topsep=0pt]
\item \textbf{TOP}: This rank locations based on the global frequencies in $S^{train}$ and \textbf{U-TOP} rank locations in $s_{u_{m}}^{train}$ based on each users' individually most frequent locations.
\item \textbf{TAXI} \cite{grabPOI}: This frequency method considers both city-wide location frequencies and users' historical destination frequencies.
\item \textbf{MF} \cite{MF}: MF is a traditional approach to many recommendation problems and can be used to learn the user-destination matrix. 

\item \textbf{RNN} \cite{rnn}: RNN takes advantage of sequential dependencies in location visit sequences with a basic recurrent structure. Its variants of \textbf{LSTM} \cite{lstm} and \textbf{GRU} \cite{gru} includes the use of different multiplicative gates to regulate information flow. For fair comparison, we extend the RNN, GRU and LSTM baselines to $f(\vec{o}_{t_{i}}\:||\:\vec{d}_{t_{i-1}} )$ where $f(.)$ is the model of choice with the concatenated inputs of the current origin $\vec{o}_{t_{i}}$ and previous destination $\vec{d}_{t_{i-1}}$ to predict the next destination $d_{t_{i}}$. This ensures the use of both origin and destination information instead of just the latter, making it the same as the OD input to our STOD-PPA's decoder in Eq. (19).


\item \textbf{HST-LSTM} \cite{hstlstm}: This LSTM based model includes spatial and temporal intervals into LSTM existing gates. Following \cite{stgcn}, we apply their proposed ST-LSTM variant here as the data does not include session information.
\item \textbf{STGN} \cite{stgcn}: A LSTM variant that learns both long and short term location visit preferences using new distance and time gates to model spatial and temporal factors, as well as a new cell state. Their variant \textbf{STGCN} removes the forget gate for better performance and efficiency.

For HST-LSTM, STGN and STGCN, these methods use the spatial and temporal intervals between $d_{t_{i-1}}$ and $d_{t_{i}}$ to predict $d_{t_{i}}$, which, as explained in Section \ref{sec:stlstm}, cannot be used in practice as this requires knowing the details of $d_{t_{i}}$ in advance. Here, we use $d_{t_{i-2}}$ and $d_{t_{i-1}}$ instead to compute the intervals, so as to leverage the most recent available historical destination visits to predict $d_{t_{i}}$.

\item \textbf{LSTPM} \cite{LSTPM}: A LSTM-based variant that models long term preferences through the use of a nonlocal network, and short term preferences via a geo-dilated network. LSTPM is the state-of-the-art approach for the next destination recommendation task.

\item \textbf{LSTPM-OD}: Same as the RNN, GRU and LSTM baselines, for fair comparison, we extend LSTPM to LSTPM-OD to use $f(\vec{o}_{t_{i}}\:||\:\vec{d}_{t_{i-1}} )$ for each timestep instead of just the previous destination $f(\vec{d}_{t_{i-1}})$ as described in their work \cite{LSTPM}.
\item \textbf{STOD-PPA} 
Our proposed approach, as shown in Fig. 4, using an encoder-decoder framework with our ST-LSTM as the encoder and the PPA as the decoder. 
\end{itemize}


\begin{table*}[t]
\caption{Performance in Acc@$K$ and MAP on seven user trajectory datasets from the transportation domain.} 
\resizebox{0.83\linewidth}{4.6cm}{
\begin{tabular}{@{}ccccccccccccccc@{}}
\toprule
 &  & \textbf{TOP} & \textbf{U-TOP} & \textbf{TAXI} & \textbf{MF} & \textbf{RNN} & \textbf{GRU} & \textbf{LSTM} & \textbf{HST-LSTM} & \textbf{STGN} & \textbf{STGCN} & \textbf{LSTPM} & \textbf{LSTPM-OD} & \textbf{STOD-PPA} \\ \midrule
{\textbf{SE-1}} & \textbf{Acc@1} & 0.0000 & 0.1344 & 0.0000 & 0.0155 & 0.1694 & 0.1621 & 0.2039 & 0.1103 & 0.0294 & 0.0530 & 0.3402$\pm$0.001 & {\ul 0.3683$\pm$0.002} & \textbf{0.4173$\pm$0.002} \\
 & \textbf{Acc@5} & 0.0127 & 0.5275 & 0.0139 & 0.0232 & 0.2340 & 0.2348 & 0.2978 & 0.2181 & 0.1012 & 0.1407 & 0.5001$\pm$0.002 & {\ul 0.5357$\pm$0.002} & \textbf{0.5615$\pm$0.001} \\
 & \textbf{Acc@10} & 0.0246 & {\ul 0.5791} & 0.0280 & 0.0380 & 0.2642 & 0.2694 & 0.3388 & 0.2737 & 0.1501 & 0.2019 & 0.5380$\pm$0.003 & 0.5689$\pm$0.002 & \textbf{0.5846$\pm$0.002} \\
 & \textbf{MAP} & 0.0147 & 0.2956 & 0.0170 & 0.0294 & 0.2039 & 0.2006 & 0.2517 & 0.1666 & 0.0724 & 0.1040 & 0.4162$\pm$0.001 & {\ul 0.4482$\pm$0.001} & \textbf{0.4865$\pm$0.002} \\ \midrule
{\textbf{SE-2}} & \textbf{Acc@1} & 0.0000 & 0.1250 & 0.0000 & 0.0164 & 0.1651 & 0.1636 & 0.2055 & 0.1051 & 0.0380 & 0.0525 & 0.3260$\pm$0.002 & {\ul 0.3617$\pm$0.002} & \textbf{0.4108$\pm$0.002} \\
 & \textbf{Acc@5} & 0.0101 & 0.5233 & 0.0115 & 0.0267 & 0.2555 & 0.2648 & 0.3300 & 0.2324 & 0.1266 & 0.1588 & 0.5064$\pm$0.001 & {\ul 0.5434$\pm$0.002} & \textbf{0.5761$\pm$0.002} \\
 & \textbf{Acc@10} & 0.0469 & 0.5814 & 0.0471 & 0.0453 & 0.2917 & 0.3104 & 0.3833 & 0.2937 & 0.1818 & 0.2270 & 0.5582$\pm$0.002 & {\ul 0.5889$\pm$0.002} & \textbf{0.6077$\pm$0.002} \\
 & \textbf{MAP} & 0.0199 & 0.2906 & 0.0209 & 0.0316 & 0.2119 & 0.2150 & 0.2678 & 0.1702 & 0.0900 & 0.1119 & 0.4119$\pm$0.001 & {\ul 0.4477$\pm$0.001} & \textbf{0.4895$\pm$0.001} \\ \midrule
{\textbf{SE-3}} & \textbf{Acc@1} & 0.0000 & 0.1285 & 0.0000 & 0.0123 & 0.1129 & 0.1003 & 0.1344 & 0.0652 & 0.0285 & 0.0384 & 0.2848$\pm$0.001 & {\ul 0.3163$\pm$0.002} & \textbf{0.3544$\pm$0.003} \\
 & \textbf{Acc@5} & 0.0156 & 0.4542 & 0.0209 & 0.0211 & 0.1771 & 0.1757 & 0.2291 & 0.1541 & 0.0741 & 0.0908 & 0.4344$\pm$0.003 & {\ul 0.4762$\pm$0.004} & \textbf{0.4985$\pm$0.001} \\
 & \textbf{Acc@10} & 0.0844 & 0.5083 & 0.0899 & 0.0381 & 0.2146 & 0.2109 & 0.2746 & 0.2048 & 0.1101 & 0.1362 & 0.4794$\pm$0.002 & {\ul 0.5176$\pm$0.004} & \textbf{0.5267$\pm$0.002} \\
 & \textbf{MAP} & 0.0274 & 0.2608 & 0.0285 & 0.0268 & 0.1489 & 0.1416 & 0.1843 & 0.1156 & 0.0621 & 0.0758 & 0.3594$\pm$0.001 & {\ul 0.3948$\pm$0.002} & \textbf{0.4249$\pm$0.002} \\ \midrule
{\textbf{SE-4}} & \textbf{Acc@1} & 0.0000 & 0.0856 & 0.0000 & 0.0140 & 0.1423 & 0.1341 & 0.1652 & 0.0830 & 0.0359 & 0.0460 & 0.2798$\pm$0.001 & {\ul 0.3047$\pm$0.001} & \textbf{0.3329$\pm$0.002} \\
 & \textbf{Acc@5} & 0.0274 & 0.4405 & 0.0274 & 0.0209 & 0.2027 & 0.2035 & 0.2558 & 0.1830 & 0.1022 & 0.1329 & 0.4222$\pm$0.002 & {\ul 0.4565$\pm$0.002} & \textbf{0.4757$\pm$0.002} \\
 & \textbf{Acc@10} & 0.0439 & 0.4912 & 0.0439 & 0.0341 & 0.2317 & 0.2363 & 0.2923 & 0.2326 & 0.1414 & 0.1870 & 0.4634$\pm$0.002 & {\ul 0.4955$\pm$0.002} & \textbf{0.5064$\pm$0.002} \\
 & \textbf{MAP} & 0.0192 & 0.2323 & 0.0200 & 0.0264 & 0.1743 & 0.1709 & 0.2116 & 0.1355 & 0.0743 & 0.0948 & 0.3492$\pm$0.001 & {\ul 0.3780$\pm$0.001} & \textbf{0.4018$\pm$0.002} \\ \midrule
{\textbf{SE-5}} & \textbf{Acc@1} & 0.0000 & 0.1063 & 0.0000 & 0.0123 & 0.1065 & 0.0940 & 0.1249 & 0.0546 & 0.0242 & 0.0286 & 0.2507$\pm$0.001 & {\ul 0.2739$\pm$0.001} & \textbf{0.3049$\pm$0.002} \\
 & \textbf{Acc@5} & 0.0679 & 0.3985 & 0.0699 & 0.0219 & 0.1603 & 0.1537 & 0.2087 & 0.1361 & 0.0858 & 0.0972 & 0.3847$\pm$0.002 & {\ul 0.4184$\pm$0.003} & \textbf{0.4433$\pm$0.003} \\
 & \textbf{Acc@10} & 0.1050 & 0.4583 & 0.1055 & 0.0365 & 0.1910 & 0.1890 & 0.2498 & 0.1818 & 0.1323 & 0.1405 & 0.4333$\pm$0.003 & {\ul 0.4617$\pm$0.003} & \textbf{0.4732$\pm$0.003} \\
 & \textbf{MAP} & 0.0392 & 0.2297 & 0.0409 & 0.0265 & 0.1385 & 0.1278 & 0.1705 & 0.1010 & 0.0638 & 0.0714 & 0.3185$\pm$0.001 & {\ul 0.3453$\pm$0.001} & \textbf{0.3726$\pm$0.002} \\ \midrule
{\textbf{SE-6}} & \textbf{Acc@1} & 0.1278 & 0.1093 & 0.1278 & 0.0078 & 0.0981 & 0.0830 & 0.1291 & 0.0653 & 0.0865 & 0.0998 & 0.2369$\pm$0.001 & {\ul 0.2568$\pm$0.001} & \textbf{0.2863$\pm$0.003} \\
 & \textbf{Acc@5} & 0.1769 & 0.3599 & 0.1775 & 0.0153 & 0.1760 & 0.1721 & 0.2405 & 0.1539 & 0.1795 & 0.1854 & 0.3942$\pm$0.004 & {\ul 0.4234$\pm$0.003} & \textbf{0.4647$\pm$0.004} \\
 & \textbf{Acc@10} & 0.1769 & 0.4411 & 0.1782 & 0.0331 & 0.2256 & 0.2130 & 0.2933 & 0.2077 & 0.2262 & 0.2366 & 0.4588$\pm$0.003 & {\ul 0.4814$\pm$0.004} & \textbf{0.5127$\pm$0.004} \\
 & \textbf{MAP} & 0.1567 & 0.2161 & 0.1574 & 0.0233 & 0.1418 & 0.1283 & 0.1876 & 0.1171 & 0.1368 & 0.1485 & 0.3161$\pm$0.001 & {\ul 0.3395$\pm$0.002} & \textbf{0.3706$\pm$0.002} \\ \midrule
{\textbf{SE-7}} & \textbf{Acc@1} & 0.0257 & 0.0719 & 0.0257 & 0.0124 & 0.0948 & 0.0869 & 0.1078 & 0.0586 & 0.0206 & 0.0313 & 0.2231$\pm$0.001 & {\ul 0.2407$\pm$0.001} & \textbf{0.2709$\pm$0.001} \\
 & \textbf{Acc@5} & 0.0838 & 0.3638 & 0.0838 & 0.0202 & 0.1318 & 0.1304 & 0.1639 & 0.1245 & 0.0563 & 0.0833 & 0.3297$\pm$0.002 & {\ul 0.3663$\pm$0.002} & \textbf{0.3895$\pm$0.001} \\
 & \textbf{Acc@10} & 0.1026 & {\ul 0.4106} & 0.1096 & 0.0305 & 0.1544 & 0.1577 & 0.1937 & 0.1614 & 0.0814 & 0.1209 & 0.3667$\pm$0.003 & 0.4012$\pm$0.002 & \textbf{0.4148$\pm$0.002} \\
 & \textbf{MAP} & 0.0521 & 0.1902 & 0.0539 & 0.0234 & 0.1166 & 0.1120 & 0.1398 & 0.0954 & 0.0455 & 0.0643 & 0.2774$\pm$0.001 & {\ul 0.3022$\pm$0.001} & \textbf{0.3289$\pm$0.001} \\ \midrule
\end{tabular}
}
\end{table*}

\begin{table*}[t]
\caption{Performance for the cold start problem in Acc@$1$.}
\resizebox{0.83\linewidth}{1.41cm}{
\begin{tabular}{@{}ccccccccccccccc@{}} \midrule
 & \textbf{TOP} & \textbf{U-TOP} & \textbf{TAXI} & \textbf{MF} & \textbf{RNN} & \textbf{GRU} & \textbf{LSTM} & \textbf{HST-LSTM} & \textbf{STGN} & \textbf{STGCN} & \textbf{LSTPM} & \textbf{LSTPM-OD} & \textbf{STOD-PPA} \\ \midrule
\textbf{SE-1} & 0.0098 & 0.0476 & 0.0098 & 0.0129 & 0.1124 & 0.1162 & 0.1377 & 0.0904 & 0.0664 & 0.0610 & 0.2416 & {\ul 0.2469} & \textbf{0.2563} \\
\textbf{SE-2} & 0.0152 & 0.0394 & 0.0152 & 0.0113 & 0.1285 & 0.1394 & 0.1599 & 0.1059 & 0.0877 & 0.1020 & 0.2235 & \textbf{0.2326} & {\ul 0.2324} \\
\textbf{SE-3} & 0.0000 & 0.0572 & 0.0000 & 0.0130 & 0.0952 & 0.1045 & 0.1246 & 0.0965 & 0.0749 & 0.0761 & 0.2319 & {\ul 0.2400} & \textbf{0.2402} \\
\textbf{SE-4} & 0.0102 & 0.0435 & 0.0102 & 0.0159 & 0.1060 & 0.1103 & 0.1320 & 0.0990 & 0.0737 & 0.0840 & 0.2045 & {\ul 0.2114} & \textbf{0.2136} \\
\textbf{SE-5} & 0.0000 & 0.0584 & 0.0000 & 0.0178 & 0.0955 & 0.1007 & 0.1219 & 0.0880 & 0.0619 & 0.0599 & 0.2260 & {\ul 0.2267} & \textbf{0.2360} \\
\textbf{SE-6} & \textbf{0.3574} & 0.0534 & \textbf{0.3574} & 0.0205 & 0.1491 & 0.1605 & 0.1648 & 0.1201 & 0.0622 & 0.0945 & 0.1836 & 0.1922 & {\ul 0.2154} \\
\textbf{SE-7} & 0.0087 & 0.0431 & 0.0090 & 0.0080 & 0.0815 & 0.0822 & 0.1029 & 0.0601 & 0.0369 & 0.0483 & 0.1760 & {\ul 0.1838} & \textbf{0.1946} \\ \midrule
\end{tabular}%
}
\end{table*}

Same as \cite{stgcn} and other existing works, we use the standard metrics of Acc@$K$ where $K\in\{1,5,10\}$ and Mean Average Precision (MAP) for evaluation. Acc@$K$ measures the performance of the recommendation set up to $K$, where the smaller $K$ is, the more challenging it is to perform well, such as Acc@1 where a score of 1 is awarded only if the ground truth next destination is in the first position ($K=1$) of the predicted ranked set, i.e. given the highest probability, and 0 score otherwise. Unlike Acc@$K$ that focuses on top $K$, MAP evaluates the quality of the entire recommendation set and measures the overall performance of the model.

\subsection{Experimental Settings}
We apply the Adam optimizer with a batch size of 1 user using cross entropy loss for the multi-classification problem, and used 15 epochs and a learning rate of 0.0001 for training. We set our location, user, timeslot, Geohash embedding dimension $dim$, and the hidden representation dimension $Hdim$ to both be 256. For the map partitioning using Geohash grid, we use Geohash precision of 5, which corresponds to Geohash cells of approximately 4.9km$\times$4.9km.
For RNN, GRU, LSTM, and MF, we use the same settings of our model where possible for fair comparison. For all other works, we use their stated recommended settings accordingly.

\subsection{Results}
We report the evaluation results of our proposed model STOD-PPA and the baselines in Table 2. For our STOD-PPA, as well as the LSTPM-OD and LSTPM baselines, we show the averaged results of 10 runs on different random seeds and with their respective standard deviations. We observe that:

\begin{itemize}[leftmargin=*,topsep=0pt]
\item STOD-PPA outperforms all the baselines, including the state-of-the-art LSTPM and its OD extension of LSTPM-OD significantly for all metrics on all seven datasets of different countries for the origin-aware next destination recommendation task.
\item LSTPM-OD is mostly the second best among the results, even when it uses both origin and destination information, the same as our best performing STOD-PPA model. This implies that trivial inclusion of origin information by concatenation is unable to exploit the underlying semantics to best learn OD relationships and perform well for the predictive task.
\item Comparing LSTPM and LSTPM-OD, where the only difference is the additional inclusion of origin information in LSTPM-OD, we can see a notable increase of performance for all metrics on all datasets consistently for LSTPM-OD. This demonstrates that origin information is indeed important and serves as a valuable source of information for the task.
\item For the other LSTM based works of STGCN, STGN, and HST-LSTM, these methods do not perform as well. We believe that these methods were unable to learn from the available historical neighbouring intervals of $d_{t_{i-2}}$ and $d_{t_{i-1}}$ to predict $d_{t_{i}}$.
\item For RNN, LSTM, and GRU baselines, although these models also use both origin and destination information in the same way of concatenation as LSTPM-OD, these baselines, however, do not consider the spatial and temporal factors when learning location-location relationships, whereas LSTPM, LSTPM-OD, and our STOD-PPA do, but in different ways.
\item For the baselines of TOP, U-TOP, TAXI and MF, these methods do not learn the sequential transitions between locations, and hence, they do not perform as well.

\end{itemize}

\subsection{Analysis of Cold Start Performance}
Same as \cite{stgcn}, we evaluate the robustness of our model for the cold start problem by first reversing our existing preprocessing, where instead of removing users with less than 10 trips on the whole dataset, we now only consider these cold start users as they have less than 10 trips. Then, on this test set, we evaluate the methods for the challenging Acc@1 metric as shown in Table 3.
For most of the datasets, our STOD-PPA has the best results, followed by mostly LSTPM-OD being the second best. Our STOD-PPA model performed slightly poorer than LSTPM-OD by 0.0002 for the SE-2 dataset and is second best for the SE-6 dataset where TAXI and TOP have the best results. We believe that TOP and TAXI methods only performed well for the SE-6 dataset because the cold start users for this country tend to be tourists taking taxi rides to popular tourist destinations on an ad-hoc basis, where user preferences are easier to predict by city-wide location frequencies. 



\subsection{Ablation Study}
In this section, we perform an ablation study to evaluate the effectiveness of our proposed STOD-PPA model on the same challenging Acc@1 metric for all datasets on their respective test sets.

\paragraph{\textbf{Spatial and Temporal Factors}} We evaluate the effectiveness of our proposed ST-LSTM by replacing all ST-LSTM modules with LSTMs, denoting this variant as OD-PPA (without considering ST factors). Notably, in Fig. 5, for all datasets, our STOD-PPA has an increase of performance over OD-PPA, demonstrating the importance of spatial and temporal factors to learn OD relationships for the recommendation task.


\paragraph{\textbf{Encoder-Decoder}}
In Fig. 6, we evaluate the performance of STOD-PPA (encoder-decoder), encoder-only, and decoder-only, where STOD-PPA has the best results by a large margin:
\begin{itemize}[leftmargin=*,topsep=0pt]
    \item For decoder-only, we replace Eq. (17) and (18) with $h^{O}_{u_{m}} = s_{u_{m}}^{train^{O}}$ and $h^{D}_{u_{m}} = s_{u_{m}}^{train^{D}}$ respectively, effectively forcing the decoder to learn from the input sequences directly where the sequential transitions and its underlying spatial-temporal factors are not considered, hence the poor results.
    \item For encoder-only, we replace Eq. (20) with $\vec y_{t_{i}} = \vec{h}^{o}_{t_{i}}\:||\:\vec{h}^{d}_{t_{i-1}}$
where $\vec{y}_{t_{i}}$ is the concatenation of the hidden state from the current origin location $\vec{h}^{o}_{t_{i}}$ and the hidden state of the previous destination location $\vec{h}^{d}_{t_{i-1}}$ from their own ST-LSTM of $\phi^{O}(.)$ and $\phi^{D}(.)$ respectively, before Eq. (21) to predict the next destination $d_{t_{i}}$. 

\end{itemize}

\paragraph{\textbf{Personalization}}
From Fig. 7, we can see that for all datasets, our STOD-PPA, where user embedding is included in the computation of attention scores for each hidden state on Eq. (19), has the best result as compared to simple addition \cite{strnn} or concatenation \cite{deepMove} of user embeddings to the hidden representations, specifically, replacing $\vec y_{t_{i}}$ in Eq. (21) to $\vec y_{t_{i}} + \vec{u}_{m}$ and $\vec y_{t_{i}}\: ||\: \vec{u}_{m}$ respectively.
\paragraph{\textbf{Sensitivity Analysis}}
From Fig. 8(a), we can observe that the model generally converges at $Hdim=64$ and maintains very similar performances towards $Hdim=256$ with a clear insensitivity trend. Similarly, in Fig. 8(b), we can see a clear insensitivity trend from epoch 10 onward, indicating that the model has converged. 
%

\pgfplotstableread[row sep=\\,col sep=&]{
    country & OD-LSTM & STOD-PPA \\
    SE-1 & 0.3918828092 & 0.4173 \\
    SE-2 & 0.3812729622 & 0.4108 \\
    SE-3 & 0.3386545116 & 0.3544 \\
    SE-4 & 0.3195385591 & 0.3329 \\
    SE-5 & 0.2948766353 & 0.3049 \\
    SE-6 & 0.2784480305 & 0.2863 \\
    SE-7 & 0.2560370814 & 0.2709 \\
    }\STdata
\begin{figure}
\begin{adjustwidth}{}{} 
\begin{tikzpicture}
    \begin{axis}[
            ybar,
            bar width=.1cm,
            width=\linewidth,
            height=2.9cm,
            legend columns=-1,
            symbolic x coords={SE-1, SE-2, SE-3, SE-4, SE-5, SE-6, SE-7},
            xtick=data,
            xtick align=inside,
            ymin=0,ymax=0.49,
            ylabel={Acc@1},
            ylabel near ticks,
            legend style={nodes={scale=0.6, transform shape}}, 
            legend image code/.code={
        \draw [/tikz/.cd,bar width=3pt,yshift=-0.2em,bar shift=0pt]
        plot coordinates {(0cm,0.8em)};
                },
                legend cell align={left}, 
        ]
        \addplot[color=red!60, fill=red!20,  thick] table[x=country,y=STOD-PPA]{\STdata};
        \addplot[color=blue!60, fill=blue!20,  thick] table[x=country,y=OD-LSTM]{\STdata};
        \legend{\normalsize{STOD-PPA},\normalsize{OD-PPA}}
    \end{axis}
\end{tikzpicture}
\end{adjustwidth}
\vspace*{-0.3cm}
\caption{Effectiveness of proposed ST-LSTM.}
\vspace*{-0.2cm}
\end{figure}
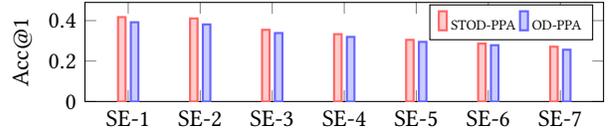

\pgfplotstableread[row sep=\\,col sep=&]{
    country & Encoder & Decoder & Encoder-Decoder \\
    SE-1 & 0.2128752813 & 0.3361352944 & 0.4173 \\
    SE-2 & 0.2100197859 & 0.3372393155 & 0.4108 \\
    SE-3 & 0.1370021384 & 0.2797679902 & 0.3544 \\
    SE-4 & 0.1659441923 & 0.2539290986 & 0.3329 \\
    SE-5 & 0.1264374458 & 0.2462462316 & 0.3049 \\
    SE-6 & 0.1311736103 & 0.2233863796 & 0.2863 \\
    SE-7 & 0.108584462 & 0.185980886 & 0.2709 \\
    }\encoderDecoder
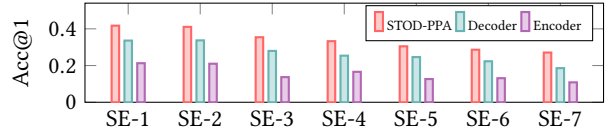
\begin{figure}
\begin{adjustwidth}{}{} 
\begin{tikzpicture}
    \begin{axis}[
            ybar,
            bar width=.1 cm,
            width=\linewidth,
            height=2.9cm,
            legend columns=-1,
            legend style={nodes={scale=0.6, transform shape}},
            symbolic x coords={SE-1, SE-2, SE-3, SE-4, SE-5, SE-6, SE-7},
            xtick=data,
            xtick align=inside,
            ymin=0,ymax=0.54,
            ylabel={Acc@1},
            ylabel near ticks,
            legend image code/.code={
        \draw [/tikz/.cd,bar width=3pt,yshift=-0.2em,bar shift=0pt]
        plot coordinates {(0cm,0.8em)};
                },
                legend cell align={left}, 
        ]
        \addplot[color=red!60, fill=red!20,  thick] table[x=country,y=Encoder-Decoder]{\encoderDecoder};
        \addplot[color=teal!60, fill=teal!20,  thick] table[x=country,y=Decoder]{\encoderDecoder};
        \addplot[color=violet!60, fill=violet!20,  thick] table[x=country,y=Encoder]{\encoderDecoder};
        \legend{\normalsize{STOD-PPA},\normalsize{Decoder},\normalsize{Encoder}}
    \end{axis}
\end{tikzpicture}
\end{adjustwidth}
\vspace*{-0.3cm}
\caption{Performance of encoder-decoder architecture.}
\vspace*{-0.2cm}
\end{figure}

\pgfplotstableread[row sep=\\,col sep=&]{
    country & Concatenate & Addition & STOD-PPA \\
    SE-1 & 0.33777548 & 0.3203196211 & 0.4173 \\
    SE-2 & 0.344237896 & 0.3415909861 & 0.4108 \\
    SE-3 & 0.2765999311 & 0.2723928807 & 0.3544 \\
    SE-4 & 0.2748186879 & 0.2704174858 & 0.3329 \\
    SE-5 & 0.2369049656 & 0.2392393416 & 0.3049 \\
    SE-6 & 0.2168444769 & 0.2130261389 & 0.2863 \\
    SE-7 & 0.2024173864 & 0.1968058938 & 0.2709 \\
    }\personalization
    
\begin{figure}

\begin{adjustwidth}{}{} 
\begin{tikzpicture}
    \begin{axis}[
            ybar,
            bar width=.1cm,
            width=\linewidth,
            height=2.9cm,
            legend columns=-1,
            legend style={nodes={scale=0.6, transform shape}},
            symbolic x coords={SE-1, SE-2, SE-3, SE-4, SE-5, SE-6, SE-7},
            xtick=data,
            xtick align=inside,
            ymin=0,ymax=0.54,
            ylabel={Acc@1},
            ylabel near ticks,
            legend image code/.code={
        \draw [/tikz/.cd,bar width=3pt,yshift=-0.2em,bar shift=0pt]
        plot coordinates {(0cm,0.8em)};
                },
                legend cell align={left}, 
        ]
        \addplot[color=red!60, fill=red!20,  thick] table[x=country,y=STOD-PPA]{\personalization};
        \addplot[color=brown!60, fill=brown!20,  thick] table[x=country,y=Concatenate]{\personalization};
        \addplot[color=darkgray!60, fill=darkgray!20,  thick] table[x=country,y=Addition]{\personalization};
        \legend{\normalsize{STOD-PPA},\normalsize{Concatenate},\normalsize{Addition}}
    \end{axis}
\end{tikzpicture}
\end{adjustwidth}
\vspace*{-0.3cm}
\caption{Comparison of personalization inclusions.}
\vspace*{-0.2cm}
\end{figure}
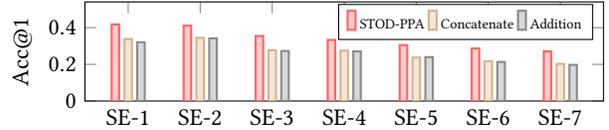

\pgfplotstableread[row sep=\\,col sep=&]{
hidden & SE-1 & SE-2 & SE-3 & SE-4 & SE-5 & SE-6 & SE-7 \\
32 & 0.3577438855 & 0.372601356 & 0.3156314061 & 0.2852567268 & 0.2678384082 & 0.2592841622 & 0.2066669076 \\
64 & 0.4051096728 & 0.4063546949 & 0.3470228139 & 0.3273708377 & 0.2982165673 & 0.2776340402 & 0.2545332891 \\
128 & 0.4099 & 0.406754022 & 0.3444450794 & 0.3263897106 & 0.3042240567 & 0.2757678217 & 0.2687990955 \\
256 & 0.4173 & 0.4108 & 0.3544 & 0.3329 & 0.3049 & 0.2863 & 0.2709 \\
    }\hidden
    
\pgfplotstableread[row sep=\\,col sep=&]{
epochs & SE-1 & SE-2 & SE-3 & SE-4 & SE-5 & SE-6 & SE-7 \\
5 & 0.3881711949 & 0.3641004313 & 0.3363672852 & 0.3147185527 & 0.2842543264 & 0.2577387624 & 0.2503180413 \\
10 & 0.4116514569 & 0.4047095289 & 0.3546144782 & 0.3269717804 & 0.3053670263 & 0.2848101622 & 0.2671321385 \\
15 & 0.4222774747 & 0.4101734259 & 0.3625278703 & 0.3344784191 & 0.3053182883 & 0.2886666725 & 0.2718396713 \\
20 & 0.4186739001 & 0.4097617032 & 0.3556588661 & 0.3322957839 & 0.3080737766 & 0.2835159517 & 0.2711319277 \\
    }\epoch
    
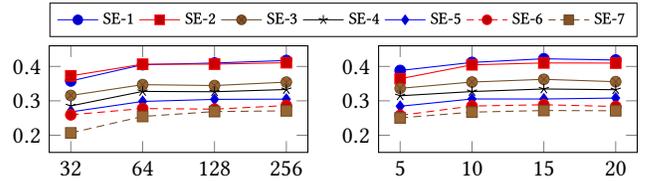
\begin{figure}[t!]
\begin{center}
\centering
\hspace*{0.458cm}
\includegraphics[width=0.925\linewidth,height=0.47cm]{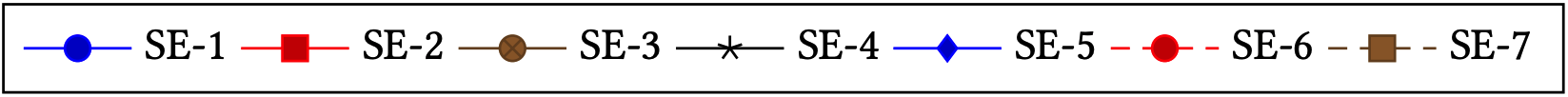}
\end{center}
    \begin{subfigure}[t]{0.5\linewidth}
    \begin{tikzpicture} 
\begin{axis}
            [
            legend columns=-1,
            width=\linewidth+0.8cm-0.5pt,
            height=3.0cm,
            symbolic x coords={32,64,128,256},
            xtick align=inside,
            ymin=0.15,ymax=0.46,
            ylabel near ticks,
                legend cell align={left}, 
        ]
        \addplot+[sharp plot] table[x=hidden,y=SE-1]{\hidden};
        \addplot+[sharp plot] table[x=hidden,y=SE-2]{\hidden};
        \addplot+[sharp plot] table[x=hidden,y=SE-3]{\hidden};
        \addplot+[sharp plot] table[x=hidden,y=SE-4]{\hidden};
        \addplot+[sharp plot] table[x=hidden,y=SE-5]{\hidden};
        \addplot+[sharp plot] table[x=hidden,y=SE-6]{\hidden};
        \addplot+[sharp plot] table[x=hidden,y=SE-7]{\hidden};

    \end{axis}
    \vspace*{-0.1cm}
\end{tikzpicture}
    \caption{Size of hidden units $Hdim$.}
  \end{subfigure}%
  ~\hspace{1pt}
  \begin{subfigure}[t]{0.5\linewidth}
     \begin{tikzpicture} 
\begin{axis}
            [
            legend columns=-1,
            legend style={at={(0.5,-0.1)},anchor=north},
            legend style={at={(0.5,-3)},
                anchor=north},
            width=\linewidth+0.8cm-0.5pt,
            height=3.0cm,
            symbolic x coords={5,10,15,20},
            xtick=data,
            xtick align=inside,
            ymin=0.15,ymax=0.46,
            ylabel near ticks,
                legend cell align={left}, 
        ]
        \addplot+[sharp plot] table[x=epochs,y=SE-1]{\epoch};
        \addplot+[sharp plot] table[x=epochs,y=SE-2]{\epoch};
        \addplot+[sharp plot] table[x=epochs,y=SE-3]{\epoch};
        \addplot+[sharp plot] table[x=epochs,y=SE-4]{\epoch};
        \addplot+[sharp plot] table[x=epochs,y=SE-5]{\epoch};
        \addplot+[sharp plot] table[x=epochs,y=SE-6]{\epoch};
        \addplot+[sharp plot] table[x=epochs,y=SE-7]{\epoch};
    \end{axis}
    \vspace*{-0.1cm}
\end{tikzpicture}
    \caption{Number of training epochs.}
    
  \end{subfigure}
  \vspace*{-0.35cm}
  \caption{Sensitivity analysis of STOD-PPA in Acc@$1$.}
    \vspace*{-0.2cm}
\end{figure}

\begin{figure}[t]
  \centering
\includegraphics[width=0.9\linewidth,height=5cm]{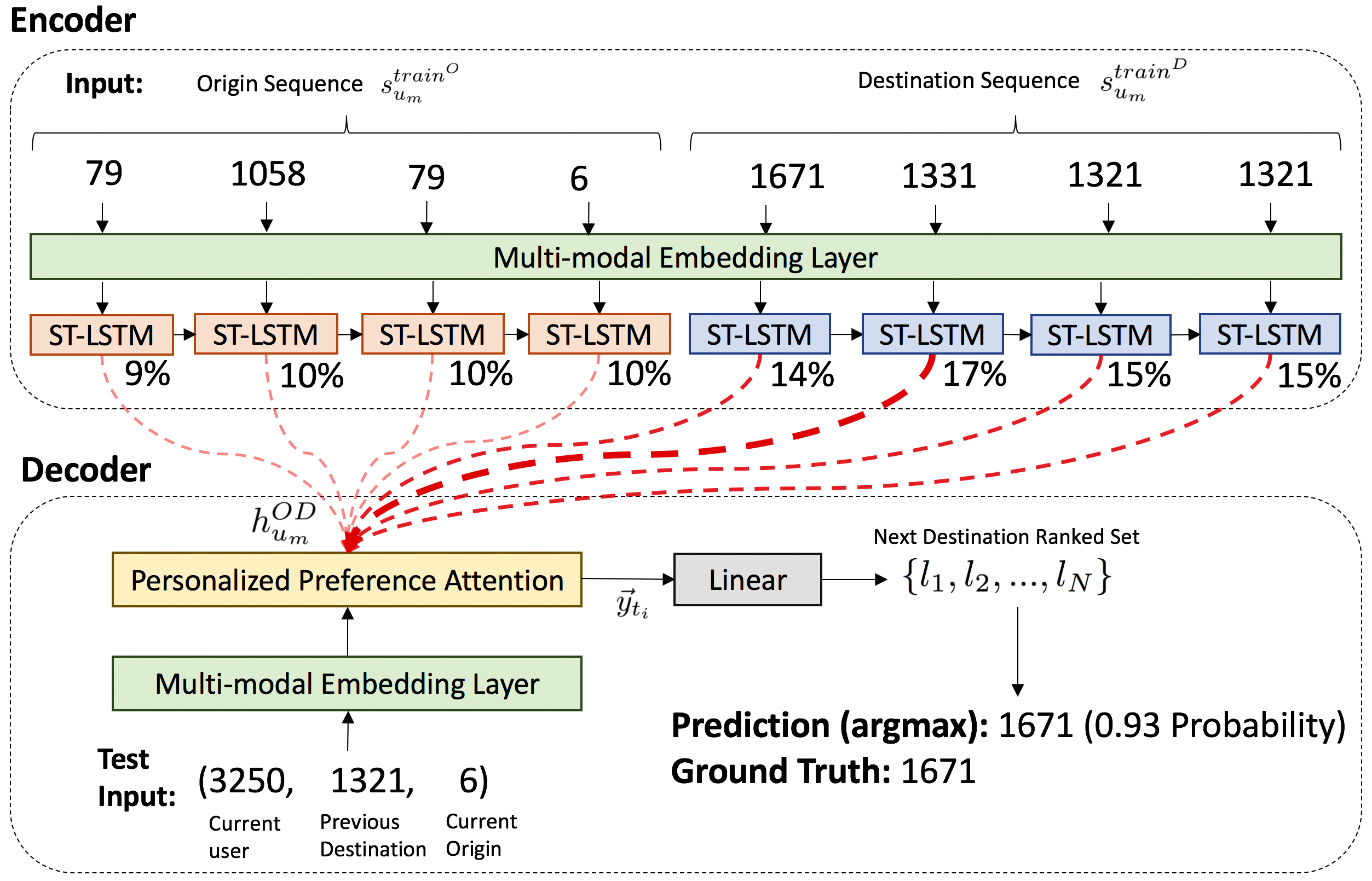}
    \caption{Interpretable user preferences with the PPA decoder on a test case from the SE-7 dataset. }
    \vspace*{-0.3cm}
\end{figure}
\subsection{Case Study: Interpretable User Preferences}
In Fig. 9, we can see a test input tuple of the user ID 3250, previous destination ID 1321 and current origin ID 6, as input to the PPA decoder where it applies the personalized preference attention on all the encoded OD hidden states from the user's historical OD sequences. In the encoder, we can see the corresponding origin and destination ID sequences, as well as the attention weights computed for each hidden state (in percentages for clarity) by the PPA decoder, where a notable difference of weights computed can be observed to best perform the predictive task and can support interpretability (e.g. transition 1671 $\rightarrow$ 1331 has the highest weight and origin ID 79 has the lowest weight). With the ground truth destination ID of 1671, our STOD-PPA approach was able to correctly predict the destination ID 1671 with the highest probability score of 0.93. 

\section{Conclusion}
This paper proposed a novel STOD-PPA encoder-decoder model for the origin-aware next destination recommendation task by learning OO, DD, and OD relationships. Specifically, we developed a ST-LSTM encoder module to allow OD relationships to be learned, as well as a PPA decoder module to learn personalized preferences among the encoded OD hidden states. Experimental results on seven real-world datasets from different countries demonstrate the effectiveness of the proposed approach with substantial improvements over existing works. For future work, we plan to study how users' side information can be used to improve the performance of this recommendation task.

\section*{Acknowledgment}
This work was funded by the Grab-NUS AI Lab, a joint collaboration between GrabTaxi Holdings Pte. Ltd. and National University of Singapore, and the Industrial Postgraduate Program (Grant: S18-1198-IPP-II) funded by the Economic Development Board of Singapore.

\bibliographystyle{ACM-Reference-Format}
\balance
\bibliography{sample-base}

\end{document}